\documentclass[final,3p,nonatbib]{elsarticle}
\usepackage[T1]{fontenc}
\usepackage{lmodern,microtype}
\usepackage{amsmath,amssymb,bm,booktabs,tabularx,graphicx}
\usepackage{enumitem,placeins,float}
\usepackage[american]{babel}
\usepackage{csquotes}
\usepackage[backend=biber,style=apa,sortcites=true]{biblatex}
\usepackage{xurl,hyperref}
\hypersetup{hidelinks,pdfauthor={Toru Takahashi, Satoru Hiwa, Tomoyuki Hiroyasu}}
\graphicspath{{figures/}}
\hypersetup{pdftitle={Matched Specialization, Different Adaptation: Second-Moment Structure in Shared Representations}}
\begin{document}
\begin{frontmatter}
\title{Matched Specialization, Different Adaptation: Second-Moment Structure in Shared Representations}
\author[dos,lod,keio,stroly]{Toru Takahashi\corref{cor}}
\ead{toru@stroly.jp}
\author[dos]{Satoru Hiwa}
\author[dos]{Tomoyuki Hiroyasu}
\cortext[cor]{Corresponding author.}
\affiliation[dos]{organization={Human Informatics and Systems Laboratory, Doshisha University},addressline={1-3 Tatara Miyakodani},city={Kyotanabe},postcode={610-0394},country={Japan}}
\affiliation[lod]{organization={Linked Open Data Initiative, NPO},addressline={2-1-2 Hitotsubashi, Chiyoda-ku},city={Tokyo},postcode={101-0003},country={Japan}}
\affiliation[keio]{organization={Keio Research Institute at SFC},addressline={5322 Endo},city={Fujisawa},postcode={252-0882},country={Japan}}
\affiliation[stroly]{organization={Stroly Inc.},addressline={109-1 Kanegaecho, Shimogyo-ku},city={Kyoto},postcode={600-8258},country={Japan}}
\begin{abstract}
When two networks reach comparable readout accuracy, can the same further training give them different gains on one signal and losses on another? We trained small networks with Adam to reconstruct two correlated Gaussian signals. A reference network consistently observed one signal more precisely. A second network placed more weight on that signal's reconstruction loss while the precisely observed signal varied across batches. After ten common updates favoring the other signal, the reference network gained more accuracy on that signal and lost more on the original. Exchanging Adam's second moments at fixed starting model weights reversed which network shifted further. An exploratory follow-up compared two batching histories at the same loss weight (0.80) on the same 40 seeds. With each network retaining its own Adam state, mixing both observation conditions within each batch produced a 0.1292 larger shift in relative readout accuracy than switching conditions between batches (95\% bootstrap interval [0.1234, 0.1344]). We then exchanged second-moment patterns between each network and the reference, matching the starting mean adaptive multiplier within each pair. Exchange reduced the reference-minus-switching contrast by 0.0985 and raised the reference-minus-mixing contrast by 0.0060, leaving the latter near zero. The reference and switching networks approached similar relative accuracy by update 500. Starting readouts also matched for three Gaussian signals but fell outside the equivalence margins for room sensors. Comparable current readouts can thus conceal differences in how optimizer history shapes the next steps of learning.

\end{abstract}
\end{frontmatter}
\section{Introduction}
\label{sec:intro}
Suppose two checkpoints favor the same signal and have comparable mean reconstruction accuracy for each of two signals. If learning now gives priority to the other signal, how much accuracy will each network gain on it, and how much will it lose on the original? This matters when choosing a checkpoint for further training. Similar evaluation scores can conceal differences under new conditions \parencite{damour2022underspecification}. Measuring both signals before and after learning shows whether one network's larger shift comes from learning more about the new priority, losing more of the old one, or both.

One way to reach the same apparent specialization is to see one signal more clearly throughout training. Another is to give its reconstruction errors more weight. We used the consistently precise network as a reference (the acquisition route). In the main comparison, the loss-weighted network learned from batches in which the precisely observed signal varied from one batch to the next (the alternating update route). These routes reached comparable mean readouts, yet the reference network both gained and lost more accuracy when learning resumed. An exploratory follow-up asked how much the grouping of observations mattered when the loss weights and the number of precise observations of each signal were held fixed.

\textcite{takahashi2026worldmodel} distinguishes two sources of disagreement. Different inference settings can yield different conclusions under the same world model. Repeated use of those settings can also change the model itself by shaping what data are encountered and how the model is updated. We vary observation precision and reconstruction-loss weights to ask whether learners with comparable readouts differ in how readily they revise their shared representations. Error-driven attention in ALCOVE illustrates how learning can favor particular stimulus dimensions \parencite{kruschke1992alcove}. Learning order can change how strongly features are represented \parencite{lampinen2024biased}. Other studies examine gradient-driven competition between features \parencite{pezeshki2021gradient} and between modality-specific encoders \parencite{huang2022modality}. A related distinction arises in models of synaptic memory, where synapses of equal strength can differ in their susceptibility to change \parencite{fusi2005cascade}. We test this distinction by exchanging retained learning state and measuring the gains and losses under the same further training.

We trained a network that compresses the two noisy signals into a two-dimensional code. Pilot runs set the loss weights for the original comparisons, and separate evaluation seeds tested five measures of readout equivalence. From each checkpoint, copies of the network received the same subsequent training. Adam retains moving averages of the gradients and their elementwise squares \parencite{kingma2015adam}. Exchanging one of these averages between networks preserves each recipient's immediate predictions and lets us measure how the transferred state changes later learning.

The main comparison tests how much of the response difference survives controls on the average adaptive step multiplier and on the distance the parameters move. We followed the acquisition and alternating routes for 500 updates to see whether their early divergence persists. In the fixed-weight follow-up, we first compared the switching and mixing networks with their own optimizer states. We then paired each network with the acquisition reference to test how exchanging second-moment patterns at a matched starting mean multiplier changes that pair's response contrast.

\FloatBarrier
\section{Retained learning state and optimizer history}
\label{sec:related}
Early input conditions influence final performance \parencite{achille2019critical}, and early SGD settings shape the subsequent optimization trajectory \parencite{jastrzebski2020breakeven}. In continual learning, networks can lose the ability to adapt to new tasks \parencite{dohare2024plasticity}. Representation-similarity metrics compare internal features at fixed endpoints \parencite{kornblith2019similarity}. Matching selected readout accuracies constrains expressed task performance without forcing internal parameters, loss geometry, or optimizer buffers to coincide; fixed-recipient state interventions isolate what retained optimizer components contribute beyond the endpoint weights.

Adaptive optimizer states are a known source of history dependence under non-stationarity. \textcite{lyle2023plasticity} link plasticity loss under task switches to loss curvature and Adam second-moment staleness; \textcite{asadi2023resetting} examine optimizer resets after objective changes; and \textcite{ellis2024local} analyze update-size transients following gradient-scale shifts. Coordinate-wise scaling also interacts with parameterization geometry \parencite{xie2025adam}, and MAdam analyzes how time-varying multi-objective weights enter Adam's second moment and reshape task-gradient interactions \parencite{liu2026madam}.

\textcite{xu2026optimizer} shows that first-moment transplantation can change later training while leaving immediate outputs unchanged. \textcite{sevetlidis2026trainingmemory} propose branching from a shared checkpoint, changing one source of training history, and comparing responses within each seed. We compare second moments from different histories with matched readouts. We separate the effects of initial mean update scale and coordinate pattern, then test whether an effect remains at matched step lengths (Table~\ref{tab:prior}).

\begin{table}[t]
\centering\small
\caption{Closest prior comparisons and the specific decomposition tested here.}
\label{tab:prior}
\begin{tabularx}{\textwidth}{>{\raggedright\arraybackslash}p{.22\textwidth}>{\raggedright\arraybackslash}X>{\raggedright\arraybackslash}X}
\toprule
Prior work & Established comparison & Present comparison\\
\midrule
\textcite{lyle2023plasticity} & Plasticity, loss geometry, and Adam instability under changing tasks & Target-specific reallocation after matching five readout measurements\\
\textcite{asadi2023resetting,ellis2024local} & Optimizer/time resets and update-size transients after objective shifts & Fixed-recipient moment swaps under global, layerwise, and movement controls\\
\textcite{xu2026optimizer} & First-moment transplantation with immediate-output invariance & Second-moment scale/pattern transplants and horizon-dependent sign changes\\
\textcite{xie2025adam,liu2026madam} & Coordinate-wise adaptivity and task-weight products in Adam's $v$ & Stationary versus alternating formation, cancellation logs, and PCGrad transplants\\
\bottomrule
\end{tabularx}
\end{table}

\section{Methods}
\label{sec:design}
\subsection{Three histories from a common starting point}
All three formation histories favor the same target. We first compare consistent precision bias with a loss-weighted route that varies which target is observed precisely across updates. A second loss-weighted route holds the mix of observation conditions fixed within each batch. It tests whether the response difference can arise without switching the precision condition between updates. Within each seed, we repeat each comparison with the target labels reversed.

Each example contains noisy observations of two correlated three-dimensional Gaussian signals, $A\sim\mathcal{N}(0,I_3)$ and $B=\rho A+\sqrt{1-\rho^2}\epsilon$ with $\epsilon\sim\mathcal{N}(0,I_3)$ and $\rho=0.35$. A seed-specific $6\times6$ orthogonal matrix $Q$ mixes the six noisy channels into one input vector. A $6\to32\to2$ encoder with $\tanh$ activations after both layers maps inputs to a two-dimensional shared code, from which two linear heads reconstruct noise-free $A$ and $B$. Both targets therefore compete for the same two-dimensional code.

Information quality and error emphasis are manipulated through distinct mechanisms. \emph{Acquisition precision} sets pre-mixing channel noise to $\sigma_{\mathrm{high}}=0.05$ (precise) or $\sigma_{\mathrm{low}}=1.20$ (imprecise), dividing each channel by $\sqrt{1+\sigma^2}$ to preserve unit input variance. \emph{Update allocation} sets explicit reconstruction-loss weights:
\begin{equation}
L_t=w_{A,t}\ell_{A,t}+w_{B,t}\ell_{B,t},\qquad w_{A,t}+w_{B,t}=1,
\label{eq:loss}
\end{equation}
where $\ell_{A,t}$ and $\ell_{B,t}$ are batch mean squared errors. Prior methods select individual features to acquire \parencite{shim2018joint}, sample from selected source tasks \parencite{chen2022active}, balance losses adaptively \parencite{chen2018gradnorm}, or project conflicting gradients \parencite{yu2020gradient}. Our comparisons ask what the resulting checkpoints retain that affects subsequent learning.

Each route starts at the same 100-update ancestor, trained with balanced observations ($\sigma=0.15$ on both targets) and equal loss weights. Formation then lasts 500 updates with coupled-$L_2$ Adam \parencite{kingma2015adam} ($\beta_1=0.9$, $\beta_2=0.999$, learning rate $0.002$). Table~\ref{tab:histories} gives the three formation histories. The term \emph{stationary} refers here to the fixed composition of each batch. We calibrate the two update routes on endpoints from 16 pilot seeds. An exploratory follow-up also uses the alternating weight for within-batch mixing. Each comparison uses all 40 evaluation seeds in its cohort (Appendix~\ref{app:measurement}).

\begin{table}[!htbp]
\centering\small
\caption{Formation histories after the shared 100-update ancestor. Each example observes one target precisely ($\sigma=0.05$) and the other imprecisely ($\sigma=1.20$). Each batch has 256 examples.}
\label{tab:histories}
\begin{tabularx}{\textwidth}{@{}p{.25\textwidth}>{\raggedright\arraybackslash}Xr@{}}
\toprule
Formation route & Which target is observed precisely? & Favored loss weight\\
\midrule
Acquisition & The favored target in every example across all updates. & $0.50$\\
Alternating update & One target per batch. Each is precise for 250 updates, in randomized order. & $0.80$\\
Within-batch update (stationary control) & 128 examples with $A$ precise and 128 with $B$ precise in every batch. & \shortstack[r]{$0.79$ calibrated\\$0.80$ follow-up}\\
\bottomrule
\end{tabularx}
\end{table}

\subsection{Measuring readout accuracy and comparing routes}
To evaluate the shared encoder independently of head weights trained under unequal loss weights, we freeze the encoder, fit ridge-regularized linear readouts ($\lambda=10^{-3}$) on 768 unused probe examples with both targets at $\sigma=0.05$, and evaluate normalized MSEs $D_A,D_B$ on 1,536 separate test examples, normalizing each target's squared error by its held-out test-set sample variance. The allocation index
\begin{equation}
s=\frac{D_B-D_A}{D_A+D_B}
\label{eq:s}
\end{equation}
lies in $[-1,1]$ and is positive when $A$ is read out more accurately than $B$. We use $D_{\mathrm{orig}}$ for the error on the originally favored target and $D_{\mathrm{dest}}$ for the error on the other target, which the opposing pulse favors. Table~\ref{tab:signs} gives the sign conventions for comparing pulse responses.

\begin{table}[!htbp]
\centering\small
\caption{Sign conventions for the two-target comparisons.}
\label{tab:signs}
\begin{tabularx}{\textwidth}{@{}p{.20\textwidth}p{.36\textwidth}X@{}}
\toprule
Quantity & Calculation & A positive value means \\
\midrule
Aligned allocation &
$s_{\mathrm{aligned}}=(D_{\mathrm{dest}}-D_{\mathrm{orig}})/(D_{\mathrm{dest}}+D_{\mathrm{orig}})$ &
The originally favored target is read out more accurately. \\
\addlinespace[2pt]
Loss of specialization &
$s_{\mathrm{aligned}}(0)-s_{\mathrm{aligned}}(t)$ &
Relative readout accuracy has shifted toward the destination target. \\
\addlinespace[2pt]
Route contrast &
\shortstack[l]{Acquisition-route loss of specialization\\minus update-route loss of specialization} &
The acquisition route shifts further toward the destination. \\
\addlinespace[2pt]
Common-scale pattern effect &
Own-pattern contrast minus exchanged-pattern contrast at matched initial mean multiplier &
Exchanging the patterns reduces the route contrast. \\
\bottomrule
\end{tabularx}
\end{table}

Because a ratio index can mask overall degradation, endpoint equivalence requires 95\% paired seed-bootstrap intervals for five measurements to lie wholly inside tolerances fixed on pilot seeds before confirmation: $\pm0.02$ for aligned $s$, $D_{\mathrm{orig}}$, $D_{\mathrm{dest}}$, and $D_A+D_B$, and $\pm0.01$ for aligned $s$ under a both-imprecise ($\sigma=1.20$) probe. As an additional check without redirection, we compare the time-averaged neutral-continuation integral $\frac{1}{200}\int_0^{200}s_{\mathrm{aligned}}(t)\,dt$ (trapezoidal rule over 200 balanced equal-weight updates). These tolerances are operational resolutions fixed prior to confirmation rather than domain-external error thresholds (Appendix~\ref{app:measurement}).

Each seed provides a shared 100-update ancestor and paired data/batch streams across compared branches. Averaging the two mirrored target directions within each seed yields 40 independent paired observations per confirmatory cohort, summarized by 95\% percentile seed-bootstrap intervals.

Once endpoint equivalence is established, identical opposing pulses test short-horizon reallocation alongside target-specific errors $D_{\mathrm{orig}}(t)$ and $D_{\mathrm{dest}}(t)$. Fixed-recipient moment transplants, start-of-pulse scale normalizations, per-step movement controls, and 500-update continuations then separate scale, coordinate pattern, formation dynamics, and horizon dependence.

\subsection{Implementation, analysis, and use of AI tools}
\label{sec:ai-methods}
The first author set the research question, selected the experimental comparisons and directed follow-up analyses. OpenAI's ChatGPT and Codex assisted with the design of controls, the implementation and execution of the Python experiments, and the aggregation and analysis of results. PyTorch performed the network updates. The analysis code computed readout errors for each target direction, averaged the two directions within each seed, and resampled seeds for the reported intervals. For the fixed-weight follow-up, the archived plan, checkpoints and directional records allow readers to repeat the training or recompute the paired estimates. The plots were rendered from experimental records with AI-assisted Python code. The records identify the training software and seeds, but do not specify a generative-model version for each coding or plotting step. The data and code availability statement identifies the versioned repository.

\section{Matched specialization conceals a response difference}
\label{sec:latent}
\subsection{Do the two histories reach a comparable starting point?}

Observation precision alone suffices to specialize the shared bottleneck: under equal loss weights, 500 updates with $A$ or $B$ precise yield mean $s=0.3385$ or $-0.3338$ across 40 confirmatory seeds, whereas alternating balanced observations keep $s$ near zero (Figure~\ref{fig:headline}a). In the matched-route cohort, calibrating the alternating update route's favored weight to $0.80$ satisfies all five pre-specified endpoint equivalence criteria simultaneously (Table~\ref{tab:endpoints}), including individual target errors around $0.40$ (favored) and $0.81$--$0.82$ (unfavored). Time-averaged neutral-continuation integrals over 200 balanced updates also agree within $\pm0.02$ under both inherited Adam ($-0.0025\;[-0.0064,0.0013]$) and fresh Adam ($-0.0018\;[-0.0068,0.0030]$; Appendix~\ref{app:optimizer}).

\begin{table}[t]
\centering\small
\caption{Confirmatory endpoint equivalence before opposing learning (40 seeds, two mirrored directions averaged per seed). The intervals for all five mean paired differences lie inside their pre-specified tolerances.}
\label{tab:endpoints}
\begin{tabular}{lcccc}
\toprule
Endpoint measurement & Acquisition route & Update route & Paired difference [95\% CI] & Tolerance\\
\midrule
Aligned allocation $s_{\mathrm{aligned}}$ & $0.3375$ & $0.3439$ & $-0.0064\;[-0.0098,-0.0032]$ & $\pm0.02$\\
Favored-target error $D_{\mathrm{orig}}$ & $0.4034$ & $0.4000$ & $+0.0034\;[+0.0014,+0.0056]$ & $\pm0.02$\\
Destination-target error $D_{\mathrm{dest}}$ & $0.8144$ & $0.8198$ & $-0.0054\;[-0.0079,-0.0029]$ & $\pm0.02$\\
Total error $D_A+D_B$ & $1.2177$ & $1.2197$ & $-0.0020\;[-0.0037,-0.0002]$ & $\pm0.02$\\
Low-precision aligned $s_{\mathrm{aligned}}$ & $0.1116$ & $0.1132$ & $-0.0016\;[-0.0028,-0.0005]$ & $\pm0.01$\\
\bottomrule
\end{tabular}
\end{table}

\subsection{Does the same new learning reveal a difference?}
We next redirect each formed checkpoint toward the previously unfavored destination target for ten updates under three opposing pulses: \emph{acquisition-only} (destination observed at $\sigma=0.05$, equal weights $0.50$), \mbox{\emph{update-only}} (alternating balanced observations, destination weight $0.80$), and \emph{combined} (destination at $\sigma=0.05$, destination weight $0.85$), comparing inherited Adam against fresh Adam (\emph{reset}).

\begin{figure}[t]
\centering
\includegraphics[width=.98\textwidth]{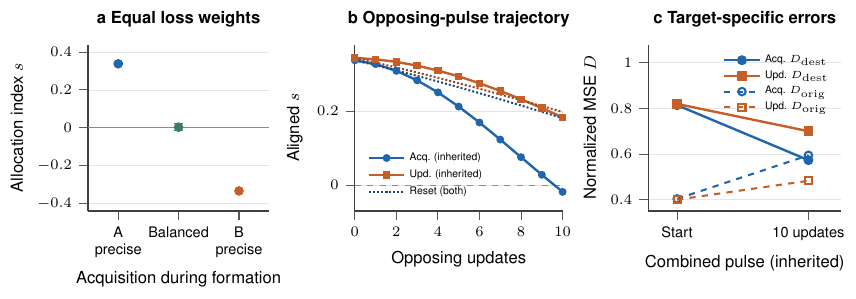}
\caption{Matched readout specialization conceals different short-horizon reallocation. (a) Observation precision alone selects specialization under equal loss weights (40 seeds). (b) Step-by-step aligned $s$ during ten combined opposing updates under inherited Adam (solid) and fresh Adam reset (dotted). (c) Target-specific readout errors under inherited Adam: faster reallocation reflects both larger error reduction on the newly favored target ($D_{\mathrm{dest}}$) and a larger error increase on the original target ($D_{\mathrm{orig}}$). Plotted with Python code developed with OpenAI tools (Section~\ref{sec:ai-methods}).}
\label{fig:headline}
\end{figure}

Under inherited Adam, acquisition-formed checkpoints reallocate substantially further than update-formed checkpoints across all three pulses, with every 95\% interval exceeding the pre-specified $+0.05$ threshold and reaching $0.1951\;[0.1880,0.2021]$ under the combined pulse (Table~\ref{tab:pulses}; Figure~\ref{fig:headline}b). Evaluated across all 80 direction-paired checkpoints (40 seeds $\times$ 2 mirrored directions), $67.5\%$ ($54/80$) of acquisition-formed checkpoints cross $s_{\mathrm{aligned}}\le0$ by update 10 (identical whether checked at update 10 or at any update $1\le t\le10$), versus $0\%$ ($0/80$) for update-formed checkpoints. Because acquisition-formed mean $s_{\mathrm{aligned}}(10)$ lands at $-0.0182$ just below zero, this binary switch rate is threshold-sensitive, and we treat the continuous route contrast as primary.

The individual target errors show a gain on one target and a cost on the other (Figure~\ref{fig:headline}c). Over ten combined updates with inherited Adam, acquisition-formed checkpoints reduce destination error $D_{\mathrm{dest}}$ from $0.8144$ to $0.5731$ while $D_{\mathrm{orig}}$ rises from $0.4034$ to $0.5943$ (lowering total error from $1.2177$ to $1.1675$), whereas update-formed checkpoints reduce $D_{\mathrm{dest}}$ only from $0.8198$ to $0.7001$ while $D_{\mathrm{orig}}$ rises from $0.4000$ to $0.4827$. Reinitializing Adam brings all three pulse contrasts inside $\pm0.02$ (Table~\ref{tab:pulses}). Reset also changes the scale of the next update. When $|g_j|$ is large relative to Adam's $\epsilon$, the first fresh update has magnitude close to $\eta$ in coordinate $j$. We therefore exchange individual moments in fixed recipients (Section~\ref{sec:mechanism}) and compare longer continuations with momentum-free SGD (Section~\ref{sec:time}).

\begin{table}[t]
\centering\small
\caption{Ten-update route contrast in loss of specialization (acquisition-formed minus update-formed decrease in $s_{\mathrm{aligned}}$; 40 confirmatory seeds). Inherited-state intervals lie above $+0.05$; reset intervals lie inside $\pm0.02$.}
\label{tab:pulses}
\begin{tabular}{lcc}
\toprule
Opposing pulse & Adam inherited & Adam reset\\
\midrule
Acquisition only & $0.1668\;[0.1601,0.1736]$ & $0.0011\;[-0.0010,0.0035]$\\
Update only & $0.1510\;[0.1456,0.1562]$ & $0.0120\;[0.0093,0.0149]$\\
Combined & $0.1951\;[0.1880,0.2021]$ & $0.0103\;[0.0079,0.0129]$\\
\bottomrule
\end{tabular}
\end{table}

Resetting Adam leaves parameter differences between the two routes intact while making ten-update responses agree within $\pm0.02$. Mean linear CKA between paired route endpoints is $0.927$ in this cohort (and $0.9575\;[0.9421,0.9707]$ in the follow-up cohort, compared with $0.6398\;[0.6180,0.6613]$ and $0.6324\;[0.6095,0.6549]$ relative to their shared 100-update ancestor), indicating high but imperfect representational alignment. We examine later response differences in Section~\ref{sec:time}.

\subsection{Does the response difference require switching between updates?}
\label{sec:stationary}
The calibrated within-batch route ($w=0.79$) meets all five endpoint criteria against the acquisition route (Appendix~\ref{app:optimizer}). On the same 40 seeds (20000--20039), the ten-update contrast under inherited Adam is $+0.0568\;[0.0514,0.0626]$ against within-batch mixing and $+0.1907$ against alternating updates. The within-batch interval just clears the $+0.05$ reference. At update ten, 62 of 80 acquisition checkpoints and 8 of 80 within-batch checkpoints reach or pass parity ($77.5\%$ and $10.0\%$).

For an exploratory follow-up, we trained the within-batch route at the alternating route's favored loss weight of $0.80$, using the same ancestors, sample indices and subsequent updates. All five starting readout measures met the equivalence criteria against both original routes (Appendix Table~\ref{tab:equalweight}). On the same 40 seeds, the inherited-state contrast against the acquisition reference was $+0.0616$ for within-batch mixing. The reference responses were identical in the two comparisons on every seed, so we can compare the two batching histories directly. The mean ten-update shift toward the new target was $0.2905$ for within-batch mixing and $0.1613$ for between-batch switching, a paired difference of $+0.1292\;[+0.1234,+0.1344]$. A descriptive comparison of the saved responses after resetting Adam gave a smaller gap of $0.0168$. Resetting both moments and the step counter together leaves open which component accounts for the change.

\section{How retained state changes adaptation}
\label{sec:mechanism}
\subsection{Which moment changes the response?}
\label{sec:moments}
Adam retains a moving average of signed gradients ($m$) and a moving average of squared gradients in each coordinate ($v$). A reset changes both, along with the step counter used for bias correction. On 40 new confirmatory seeds we instead replace one moment at a time with the corresponding buffer from the paired checkpoint. Both checkpoints descend from the same ancestor, so their coordinates correspond. Each recipient keeps its weights, its other moment and its step count ($t=600$). Its initial readouts are identical with either donor.

\begin{figure}[t]
\centering
\includegraphics[width=.98\textwidth]{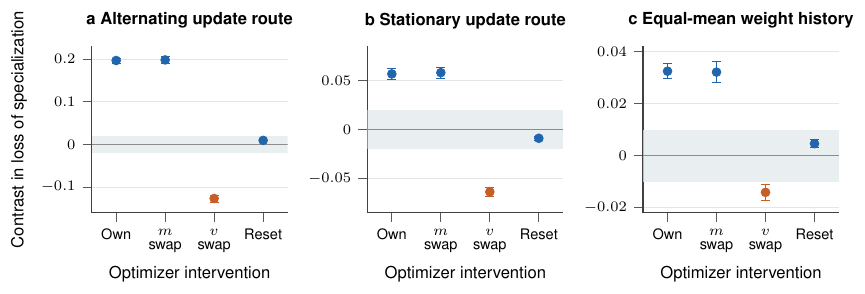}
\caption{Fixed-recipient second-moment exchange reverses short-horizon reallocation contrasts across three comparisons (40 paired seeds each). The shaded bands indicate the equivalence margins used in the original protocols and as references for the within-batch control. (a) Acquisition route minus alternating update route ($\pm0.02$). (b) Acquisition route minus the calibrated within-batch route ($w=0.79$) ($\pm0.02$). (c) Constant minus variable loss-weight history at fixed acquisition and equal mean weight ($\pm0.01$). Plotted with Python code developed with OpenAI tools (Section~\ref{sec:ai-methods}).}
\label{fig:moments}
\end{figure}

Exchanging $m$ leaves both recipients' mean ten-update responses close to their own-state values. The change in the alternating route contrast falls within the pre-specified $\pm0.02$ band. Exchanging $v$ reverses the contrast from $+0.1968$ to $-0.1265$ in this cohort (Figure~\ref{fig:moments}a and Appendix~Table~\ref{tab:components}). Both target-specific error gaps also reverse under $v$ exchange in the 40-seed follow-up cohort (Appendix~\ref{app:optimizer}). Against the within-batch route, the exchange likewise reverses the contrast, and reset brings it inside $\pm0.02$ (Figure~\ref{fig:moments}b).

Adam accumulates squared weighted gradients, including cross-task products \parencite{liu2026madam}. Holding observations fixed, we compare a constant favored weight of $0.80$ with a schedule that uses $1.00$ and $0.60$ for 250 updates each in random order. The two histories have the same mean weight and matched readouts on 40 separate seeds. The constant-weight history loses more specialization over ten updates. Exchanging $v$ reverses the contrast from $+0.0324$ to $-0.0141$, while resetting Adam brings it within $\pm0.01$ (Figure~\ref{fig:moments}c and Appendix~\ref{app:optimizer}).

\subsection{Does matching the initial update scale remove the exchange effect?}
\label{sec:scale}
Acquisition checkpoints start the pulse with a larger mean adaptive step multiplier $\bar a_v=P^{-1}\sum_j\eta/(\sqrt{\hat v_j}+\epsilon)$ in both comparisons. On the same 40 follow-up seeds, we rescale $v$ at pulse start to match $\bar a_v$ across routes while keeping its relative pattern across parameter coordinates. We also replace $v$ with a scalar in each route, once preserving its own mean multiplier and once using the common mean multiplier. The recipient keeps its weights, $m$, and step count ($t=600$). Its moments evolve normally during the pulse (Appendix~\ref{app:optimizer}).

\begin{figure}[t]
\centering
\includegraphics[width=.98\textwidth]{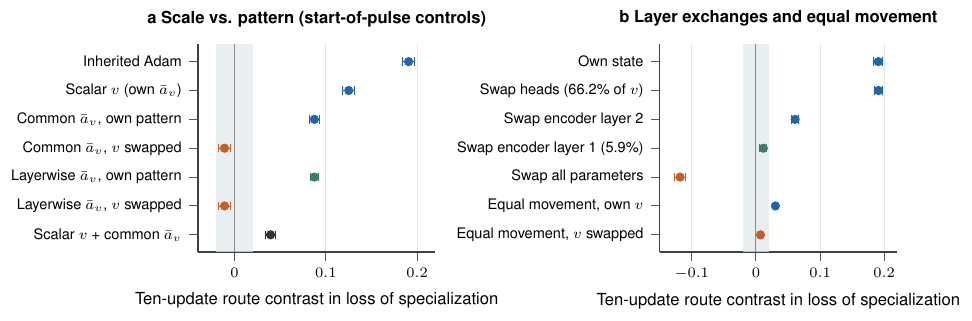}
\caption{Ten-update response differences in the alternating comparison (40 paired seeds). Shading marks $\pm0.02$. (a) Exchanging $v$ changes the contrast by about $0.098$ at a common starting mean multiplier, whether the means are matched globally or within each layer. (b) Exchanging the first encoder layer alone reduces the positive contrast from $0.1907$ to $0.0115$. Exchanging all layers reverses it to $-0.1183$. The bottom pair shows responses under steps of equal length. Appendix Table~\ref{tab:ivfmechanism} gives the values of all plotted points. Plotted with Python code developed with OpenAI tools (Section~\ref{sec:ai-methods}).}
\label{fig:scalepattern}
\end{figure}

Matching the initial global multiplier removes the positive own-state contrast against the within-batch route. At $w=0.79$ the within-batch contrast moves from $-0.0065$ to $-0.0004$. The own-minus-exchanged contrast is $+0.0985$ against the alternating route and $-0.0060$ against the calibrated within-batch route. Matching the initial mean multipliers within layers gives a similar result in both comparisons, including a positive effect of $+0.0984$ for alternating updates. In the alternating comparison, the effect also remains positive at matched scales for individual weight or bias tensors (Figure~\ref{fig:scalepattern}a and Appendix~Table~\ref{tab:ivfmechanism}).

At $w=0.80$, the pattern effect against within-batch mixing again rounds to $-0.0060$. The mean paired difference from $w=0.79$ is $+0.0001$. At the common weight of $0.80$, the pattern effect against alternating updates is $+0.1045\;[+0.0987,+0.1102]$ larger on the same 40 seeds. Each acquisition/update pair uses its own common starting multiplier (Appendix~\ref{app:optimizer}).

Giving the two routes the same scalar $v$ at restart leaves a positive alternating-route contrast. It persists with both initial first moments set to zero, so the trained models also contribute to the response difference (Appendix~Table~\ref{tab:ivfmechanism}). A later difference emerges after a standard reset (Section~\ref{sec:time}). Whether the same features of the trained models account for both differences remains open.

\subsection{Where does the pattern act, and is total movement sufficient?}
Matching the mean multiplier at pulse start still allows later update lengths to differ. We run four branches together (both routes with own or swapped $v$). At every step we rescale each branch's update to the geometric mean of the four proposed Euclidean norms and keep its proposed direction (Appendix~\ref{app:layers}).

With each checkpoint's own $v$, matching global step length reduces the route contrast from $0.1907$ to $0.0303$, an $84\%$ reduction under the modified update rule. Exchanging $v$ further reduces the contrast at that matched length. Equalizing the step length separately in each layer also leaves an exchange effect of $0.0238\;[0.0219,0.0256]$. Thus the distance moved at each update alone cannot explain the exchange effect (Appendix~Table~\ref{tab:ivfmechanism}).

A separate test exchanges the buffer one layer at a time at its original scale. Exchanging only the first encoder layer's $v$ reduces the contrast from $0.1907$ to $0.0115$, even though that layer holds just $5.9\%$ of the acquisition-route buffer. Exchanging both heads leaves the contrast close to its own-state value (Figure~\ref{fig:scalepattern}b). These layer-specific exchanges change both the scale and pattern of the selected buffer. In a separate exchange of the full buffer, the effect also persists when matching the initial mean multipliers within rows or columns of the first layer (Appendix~Table~\ref{tab:ivfmechanism}).

\subsection{How do second-moment differences form during learning?}
\label{sec:forming}
Adam accumulates squared combined gradients. A small second moment could come from small gradients for each task or from substantial task gradients that oppose each other. The formation records distinguish these possibilities along the realized trajectories by comparing the squared individual contributions with the cross term of their weighted sum. Cancellation is strongest when the weighted gradients are nearly opposite and similar in magnitude.

Replaying the two original routes reproduces all 160 saved formation endpoints bitwise in the recorded environment. We record the same weighted-gradient terms during formation in the within-batch follow-up (Appendix~\ref{app:formationlogs}). The sum of squared individual gradient contributions is larger under acquisition than under alternating updates. In the first layer, the cross term cancels about $94\%$ of the sum in the acquisition route and $28\%$ in the alternating route. Cancellation reaches about $65\%$ in the calibrated within-batch route. The $94\%$ and $65\%$ rates differ even though both routes keep their observation regime fixed across updates. Among alternating batches, cancellation is especially weak when the unfavored target is observed precisely. We next test how changing gradient conflict during formation affects later learning.

\subsection{Can altering gradient conflict during formation change later response?}
Formation logs describe the gradients along the trajectories the models followed. To test the effect of altering gradient conflict, we apply two-task PCGrad projection \parencite{yu2020gradient} to conflicting weighted encoder gradients during acquisition formation on 40 new seeds (38000--38039). Then we cross native and projected-history $v$ with native and projected acquisition recipients in a $2\times2$ transplant. Each recipient keeps its starting weights, $m$, and step count ($t=600$), so we can measure how the donor $v$ changes its response (Appendix~\ref{app:projection}).

\begin{figure}[t]\centering
\includegraphics[width=.96\textwidth]{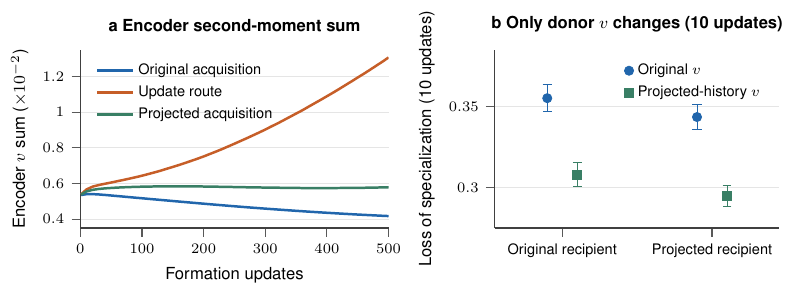}
\caption{Altering encoder gradient conflict during formation changes a fixed recipient's short-horizon response (40 seeds). (a) Mean encoder second-moment sum during 500 formation updates. (b) Ten-update loss of specialization in $2\times2$ fixed-recipient $v$ transplants (bars: 95\% paired seed-bootstrap intervals). Plotted with Python code developed with OpenAI tools (Section~\ref{sec:ai-methods}).}
\label{fig:projection}
\end{figure}

PCGrad shrinks individual encoder gradients ($0.117\times$ in squared sum) but removes opposing components so that combined squared encoder gradients grow ($1.844\times$), raising final encoder $v$ by $1.423\;[1.386,1.459]$ and lowering $\bar a_v$ to $0.713\;[0.705,0.721]$ of native acquisition (Figure~\ref{fig:projection}a). Transplanting only this projected-history $v$ into the native acquisition recipient changes ten-update loss of specialization by $-0.0472\;[-0.0489,-0.0455]$ (and by $-0.0487\;[-0.0503,-0.0470]$ in the projected recipient; Figure~\ref{fig:projection}b and Appendix~Table~\ref{tab:projection}).

Because PCGrad alters gradient magnitude, direction, and endpoint weights together (failing three endpoint equivalence bands), this fixed-recipient test isolates the causal effect of the resulting $v$ rather than a pure cancellation fraction in natural training. Moreover, rescaling projected-history $v$ to match the native recipient's starting global $\bar a_v$ (or encoder $\bar a_v$) changes the ten-update donor effect to $+0.0113\;[0.0102,0.0123]$ ($+0.0115\;[0.0104,0.0125]$), and by 50 updates even unscaled projected-history $v$ slightly increases response ($+0.0029\;[0.0025,0.0032]$). Thus the slowdown is sensitive to the starting adaptive multiplier. Under either normalization the donor still affects the response, with the opposite sign.

\section{How time and learning conditions change the effect}
\label{sec:time}
\subsection{Does early divergence persist under common learning?}
To separate transient optimizer-mediated speed differences from delayed model-weight effects and long-run convergence, we continue the acquisition and alternating update routes under the combined opposing condition for 500 updates on the 40 follow-up seeds under inherited Adam, fresh Adam (reset), and momentum-free SGD with a per-seed, per-direction learning rate calibrated from first-batch update-to-gradient norm ratios without using response outcomes (Appendix~\ref{app:optimizer}).

\begin{figure}[t]
\centering
\includegraphics[width=.98\textwidth]{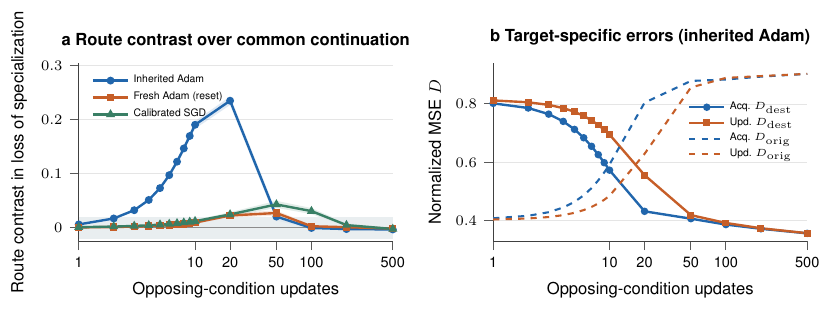}
\caption{Short-term optimizer effects and later model-side differences during continued learning (40 follow-up seeds; shaded band in (a) marks the ten-update $\pm0.02$ reference). (a) Route contrast over 500 opposing updates under inherited Adam, fresh Adam, and calibrated SGD. (b) Mean destination ($D_{\mathrm{dest}}$) and original-target ($D_{\mathrm{orig}}$) readout errors under inherited Adam. Plotted with Python code developed with OpenAI tools (Section~\ref{sec:ai-methods}).}
\label{fig:timecourse}
\end{figure}

Under inherited Adam the route contrast peaks at 10--20 updates and then shrinks (Figure~\ref{fig:timecourse}a). Target-specific error curves show why (Figure~\ref{fig:timecourse}b): acquisition-formed $D_{\mathrm{dest}}$ drops below $D_{\mathrm{orig}}$ before update 10, whereas update-formed checkpoints reach the same crossover near update 20 and catch up by update 50--100. Conversely, fresh Adam and calibrated SGD show small contrasts at update 10 ($0.0096$ and $0.0123$) but develop clear positive contrasts by update 50 ($0.0274\;[0.0227,0.0321]$ and $0.0433\;[0.0368,0.0499]$), showing that differences between the trained models remain detectable after inherited optimizer state is removed.

By update 500, the route contrasts are close to zero under all three optimizers, with inherited-Adam endpoint allocation difference reaching $-0.00045\;[-0.00146,0.00057]$ ($D_{\mathrm{dest}}\approx0.357$, $D_{\mathrm{orig}}\approx0.903$ in both routes) and inherited-minus-fresh contrast changing sign by update 50. Here inherited state changes the path to similar relative readout accuracy over the observed 500 updates.

\subsection{Changing network and formation conditions}
We test nine network and formation-regime variants on 32 paired seeds, varying hidden width ($64$), activation (ReLU), target correlation ($\rho=0$), second-moment decay $\beta_2\in\{0.9,0.99,0.999\}$ (effective memory timescale $1/(1-\beta_2)\in\{10,100,1000\}$ updates), and formation length ($500$ vs.\ $1,500$ updates) under the alternating update design. All nine alternating-route settings satisfy all five endpoint tolerances and show a positive common-scale pattern effect at matched starting $\bar a_v$ (Table~\ref{tab:claims}; Appendix~Table~\ref{tab:ivfbeta}).

Of the six decay-by-duration combinations, unnormalized $v$ exchange reverses the contrast at $\beta_2=0.999$ after both 500 and 1,500 formation updates, and at $\beta_2=0.99$ after 500. The contrast stays positive at $\beta_2=0.9$ at both durations and at $\beta_2=0.99$ after 1,500. At $\beta_2=0.9$, resetting Adam increases the contrast ($0.0373$ vs.\ $0.0298$ after 500 updates and $0.1073$ vs.\ $0.0764$ after 1,500). At all three decay rates the reset contrast exceeds $0.02$ after 1,500 updates. Longer formation thus leaves a larger difference between the trained models under reset, even where the acquisition/update $\bar a_v$ ratio shrinks (Appendix~\ref{app:extension}).

\begin{table}[!htbp]
\centering\small
\caption{Summary of ten-update claims across the two within-batch weights and 13 extension settings (Y: satisfied; N: interval establishes failure; ?: interval crosses boundary). The pattern criterion requires a positive 95\% interval for own minus swapped $v$ at a common initial $\bar a_v$. Reversal requires a positive own-state interval and a negative exchanged-state interval. Reset equivalence requires the interval to fall inside $\pm0.02$.}
\label{tab:claims}
\begin{tabular}{lcccc}
\toprule
Setting & Endpoints & \shortstack{Common-scale\\pattern effect} & Reversal & \shortstack{Reset\\equivalence}\\
\midrule
\multicolumn{5}{l}{\emph{Two-target Gaussian: stationarity control (40 seeds)}}\\
Within-batch, calibrated ($w=0.79$) & Y & N & Y & Y\\
Within-batch, equal-weight follow-up ($w=0.80$) & Y & N & Y & Y\\
\addlinespace[2pt]
\multicolumn{5}{l}{\emph{Two-target Gaussian: alternating-route variants (32 seeds)}}\\
Baseline ($\beta_2=0.999$, 500) & Y & Y & Y & Y\\
Width 64 & Y & Y & Y & N\\
ReLU & Y & Y & Y & Y\\
Independent targets ($\rho=0$) & Y & Y & Y & Y\\
$\beta_2=0.9$, 500 & Y & Y & N & N\\
$\beta_2=0.99$, 500 & Y & Y & Y & Y\\
$\beta_2=0.9$, 1500 & Y & Y & N & N\\
$\beta_2=0.99$, 1500 & Y & Y & N & N\\
$\beta_2=0.999$, 1500 & Y & Y & Y & N\\
\addlinespace[2pt]
\multicolumn{5}{l}{\emph{Three-target Gaussian (40 seeds)}}\\
Bottleneck $d_Z=2$ & Y & Y & Y & Y\\
Bottleneck $d_Z=3$ & Y & Y & Y & ?\\
\addlinespace[2pt]
\multicolumn{5}{l}{\emph{Three-target UCI room sensors (40 seeds; unmatched endpoints)}}\\
Bottleneck $d_Z=2$ & N & Y & Y & Y\\
Bottleneck $d_Z=3$ & N & Y & ? & Y\\
\bottomrule
\end{tabular}
\end{table}

\subsection{Changing the number of targets and the input data}
We add a third symmetric three-dimensional Gaussian target and use a $9\to32\to d_Z$ encoder with $d_Z\in\{2,3\}$. We test all six ordered transitions at the pilot-calibrated weight of $0.65$ on 40 confirmatory seeds. Both capacities pass all eight endpoint criteria. Exchanging $v$ reverses the ten-update contrast from $0.1945$ to $-0.1005$ at $d_Z=2$ and from $0.4054$ to $-0.2193$ at $d_Z=3$. The common-scale pattern effects are positive at ten updates ($0.0825$ and $0.1541$) and negative at 50 (Appendix~\ref{app:threeadapt}). The reset interval lies within the equivalence band at $d_Z=2$ and straddles $+0.02$ at $d_Z=3$ (Table~\ref{tab:claims} and Appendix~\ref{app:threeadapt}).

Replacing Gaussian signals with temperature and humidity pairs from three rooms in the UCI Appliances Energy Prediction dataset \parencite{candanedo2017data,candanedo2017dataset} tests unequal empirical distributions and chronological train/probe/test splits ($6\to32\to d_Z$, 40 seeds; Appendix~\ref{app:natural}). Both capacities retain positive common-scale pattern effects ($0.0103$ and $0.0257$), but pilot and confirmatory runs fail room-specific endpoint matching ($1/24$ and $0/24$ criteria passed; pilot discrepancies $5.30$--$12.44\times$ tolerance). The tested grid of scalar loss weights did not produce a match for these room-specific measurements. Matched-readout claims therefore rest on the Gaussian tasks, while the sensor task provides supplementary evidence that donor $v$ alters fixed-recipient responses on empirical inputs.

\FloatBarrier
\section{Discussion: learning history and representational revision}
\label{sec:scope}
With comparable mean readouts, the same further learning yields different gains and losses. Over ten updates the acquisition route gains more accuracy on the new target and loses more on the original than the alternating route. Second-moment exchange reverses the response contrast in both comparisons. At a common starting mean multiplier, the exchange effect is positive against the alternating route and small and negative against the within-batch route. In the alternating comparison, matching total movement sharply reduces the contrast, and matching movement within each layer leaves a small exchange effect ($0.0238$).

The fixed-weight follow-up separates two questions about the batching histories. When each network kept its own Adam state, the within-batch network shifted further toward the new target than the alternating network. In separate exchanges with the acquisition reference, the two histories also differed in how sensitive the response contrast was to the arrangement of second moments across parameters. One possible explanation is that batches containing a single precision condition reinforce different gradient coordinates on successive updates. In the alternating route, gradients cancelled less when the unfavored target was observed precisely. An intervention on those formation updates could test whether they create the coordinates that influence the later response.

Retaining Adam's state gives the largest contrast at 10--20 updates. Resetting Adam or switching to SGD reveals smaller differences near update 50. After 500 common updates, the acquisition and alternating routes have similar relative readout accuracy. Sign reversal varies across the six decay and duration combinations, leaving the transition unresolved (Appendix~Table~\ref{tab:ivfbeta}).

The state exchanges rely on corresponding parameter coordinates inherited from a shared ancestor. The matches concern selected mean linear readouts of small networks trained with coupled-$L_2$ Adam. They hold with two and three Gaussian targets. With room-sensor inputs, the within-recipient exchange is still informative, but the initial readouts differ between histories. The formation replays account for the gradients along observed paths. PCGrad provides a further intervention on those paths, changing gradient magnitude and direction together. The resulting buffer changes a fixed recipient's response, and matching its starting adaptive scale reverses the ten-update donor effect.

Our results distinguish what learners currently represent from how readily they revise it. A moment exchange preserves each recipient's immediate readouts while changing its gains and losses over the next ten updates. Synaptic memory models draw a related distinction between strength and plasticity \parencite{fusi2005cascade}. The initial gap between the acquisition and alternating routes narrows with continued learning. For research on persistent cognitive orientations, the next question is which ways of distinguishing and relating aspects of experience continue to guide learning across situations and over time.

Short continuations with inherited and reset optimizer states could measure a checkpoint's gains on a new target and losses on the original. Larger models and tasks would test whether those measurements predict downstream performance.

\section{Conclusion}
Networks with comparable current readouts differed in how much they gained on the newly favored signal and lost on the original. In the main comparison, exchanging Adam's second moments reversed which network shifted further. Exchange in the first encoder layer alone substantially reduced the contrast, and a small effect remained when movement was matched within each layer. The reference and switching histories reached similar relative accuracy by update 500. In the exploratory fixed-weight comparison, the within-batch network shifted further than the switching network with their own states. At matched starting mean multipliers within each pair, exchanging patterns with the acquisition reference sharply reduced its contrast with switching and slightly increased its near-zero contrast with mixing. Thus training histories with comparable readouts can differ both in the response they produce and in how that response depends on the optimizer state they leave behind.

\section*{Data and code availability}
The formation and continuation records, including both within-batch controls, are available at \url{https://github.com/culturegeek/matched-specialization-different-adaptation/releases/tag/v1.6.0}. The fixed-weight comparison is in \mbox{\texttt{equal-weight-080/}} under \texttt{experiments/iv-g/}, with the plan, training and verification scripts, checkpoints and results. Code is distributed under the MIT license and generated data under CC BY 4.0. The UCI sensor data retain their original CC BY 4.0 attribution.

\section*{Funding}
This research did not receive any specific grant from funding agencies in the public, commercial, or not-for-profit sectors.

\section*{CRediT authorship contribution statement}
\textbf{Toru Takahashi:} Conceptualization, Methodology, Investigation, Writing -- review and editing. \textbf{Satoru Hiwa and Tomoyuki Hiroyasu:} Writing -- review and editing. Both coauthors contributed to interpreting the comparisons in relation to the research questions and prior work, and helped shape the conclusions.

\section*{Declaration of generative AI and AI-assisted technologies in the manuscript preparation process}
The authors used ChatGPT and Codex (OpenAI) to assist with literature searches and citation checks, draft and translate passages, and revise the manuscript. They also considered comments on the manuscript produced by Claude (Anthropic) and Gemini (Google). AI-assisted plotting code produced the figures from recorded experimental results. The authors evaluated the scientific claims and revised the interpretation and wording. They take responsibility for the content, accuracy and integrity of the article.

\printbibliography

\FloatBarrier
\section*{Appendices}
\appendix
\renewcommand{\thesection}{\Alph{section}}
\counterwithin{table}{section}
\counterwithin{figure}{section}
\counterwithin{equation}{section}
\section{Measurement and paired comparisons}
\label{app:measurement}
The endpoint criteria test whether selected linear-readout measurements agree within pre-specified operational tolerances before opposing learning begins.

The central network is $6\to32\to2$ with $\tanh$ after both encoder layers and two linear heads ($308$ parameters total: $32\times6+32$ in encoder layer 1, $2\times32+2$ in encoder layer 2, and $2\times(3\times2+3)$ in the heads). Training/probe/test sample counts are $3072/768/1536$, batch size 256, Adam learning rate $0.002$, $(\beta_1,\beta_2)=(0.9,0.999)$, $\epsilon=10^{-8}$, coupled $L_2$ weight decay $10^{-5}$ \parencite{kingma2015adam} (rather than decoupled AdamW), and readout ridge coefficient $10^{-3}$. Each seed supplies a 100-update balanced ancestor ($\sigma=0.15$, equal weights $0.50$) and paired data/batch streams. Fresh linear readouts fit on unused probe data and evaluate on separate test data, normalizing each target's test MSE by its held-out test-set sample variance.

Formation endpoint tolerances are $\pm0.02$ for aligned $s$, favored-target error $D_{\mathrm{orig}}$, destination-target error $D_{\mathrm{dest}}$, and total error $D_A+D_B$, and $\pm0.01$ for low-precision aligned $s$ ($\sigma=1.20$ on both channels). Response bands are $\pm0.02$ for neutral continuation and reset contrasts, $+0.05$ for inherited-state opposing pulses, $\pm0.02$ for $m$-exchange change, and $\pm0.01$ for the within-route weight-history reset contrast.

A paired contrast is classified as equivalent when its 95\% seed-bootstrap interval lies wholly inside the pre-specified band, exceeding the pre-specified band when it lies wholly beyond either boundary, and unresolved otherwise. All three inherited-state pulse intervals lie wholly above $+0.05$ (Table~\ref{tab:pulses}), and all reset-pulse intervals lie wholly inside $\pm0.02$.

The matched-response study uses pilot seeds 8000--8015 and evaluates seeds 8100--8139. The component and equal-mean weight comparisons use separate pilot and evaluation cohorts (9000--9015 and 9100--9139, and 9200--9215 and 9300--9339, respectively). In the calibrated within-batch follow-up, seeds 8000--8015 calibrate $w=0.79$ by endpoint measurements, and seeds 20000--20039 evaluate the comparison against the saved acquisition checkpoints. The original response bands serve as references. The exploratory fixed-weight follow-up reuses those 40 evaluation seeds at $w=0.80$.

Endpoint equivalence applies to mean paired readout measurements rather than point-by-point identity of network functions (paired endpoint CKA averages $0.927$). Leave-one-seed-out linear regressions predicting ten-update response contrasts from residual endpoint differences in $D_{\mathrm{orig}}$, $D_{\mathrm{dest}}$, and low-precision aligned $s$ give MSE ratios relative to a training-mean predictor of $1.118$, $1.023$, and $0.984$ across three independent 40-seed cohorts, so these residual readout differences offer little improvement over prediction by the training mean.

\section{State interventions, stationarity control, and continuation}
\label{app:optimizer}
This appendix reports recipient-level estimates for component exchanges, endpoint and response details for the stationary within-batch control, start-of-pulse scale transformations, and 500-update continuations.

In the matched-response cohort (seeds 8100--8139; Table~\ref{tab:endpoints}), time-averaged neutral-continuation integrals $\frac{1}{200}\int_0^{200}s_{\mathrm{aligned}}(t)\,dt$ (computed by trapezoidal integration divided by 200 updates) differ by $-0.0025\;[-0.0064,0.0013]$ under inherited Adam and $-0.0018\;[-0.0068,0.0030]$ after reset. Inherited-minus-reset interactions for acquisition-only, update-only, and combined pulses are $0.1657$, $0.1390$, and $0.1848$, each exceeding the pre-specified $0.08$ gate.

In the component-exchange cohort (seeds 9100--9139), the own-minus-$m$-exchange contrast shift is $-0.0014\;[-0.0044,0.0018]$ (inside $\pm0.02$). Acquisition-formed total $v$ is $0.5577\times$ the alternating update route's, while mean inverse scaling $\bar a_v$ is $1.950\times$ larger. Table~\ref{tab:components} reports the alternating comparison's recipient responses and the contrasts for both designs. The alternating recipient means change little under $m$ exchange.

\begin{table}[!htbp]
\centering\small
\caption{Ten-update combined-pulse loss of specialization by recipient and route contrast across component interventions (40 paired seeds per comparison).}
\label{tab:components}
\begin{tabular}{lcccc}
\toprule
Condition & Acq.\ recipient (alt.) & Upd.\ recipient (alt.) & Alternating contrast & Within-batch ($w=0.79$)\\
\midrule
Own state & \shortstack{$0.3616$\\$[0.3545,0.3689]$} & \shortstack{$0.1648$\\$[0.1583,0.1712]$} & \shortstack{$0.1968$\\$[0.1910,0.2029]$} & \shortstack{$0.0568$\\$[0.0514,0.0626]$}\\
\addlinespace[3pt]
Exchange $m$ & \shortstack{$0.3620$\\$[0.3498,0.3751]$} & \shortstack{$0.1638$\\$[0.1587,0.1691]$} & \shortstack{$0.1982$\\$[0.1912,0.2054]$} & \shortstack{$0.0581$\\$[0.0526,0.0638]$}\\
\addlinespace[3pt]
Exchange $v$ & \shortstack{$0.1950$\\$[0.1909,0.1990]$} & \shortstack{$0.3215$\\$[0.3082,0.3344]$} & \shortstack{$-0.1265$\\$[-0.1348,-0.1183]$} & \shortstack{$-0.0637$\\$[-0.0683,-0.0589]$}\\
\addlinespace[3pt]
Reset & \shortstack{$0.1570$\\$[0.1541,0.1599]$} & \shortstack{$0.1473$\\$[0.1434,0.1514]$} & \shortstack{$0.0097$\\$[0.0078,0.0117]$} & \shortstack{$-0.0089$\\$[-0.0111,-0.0066]$}\\
\bottomrule
\end{tabular}
\end{table}

In the stationary within-batch update control ($w=0.79$, seeds 20000--20039), all five endpoint differences pass equivalence: aligned $s$ $-0.0059\;[-0.0084,-0.0034]$, $D_{\mathrm{orig}}$ $+0.0052\;[+0.0039,+0.0063]$, $D_{\mathrm{dest}}$ $-0.0003\;[-0.0030,+0.0023]$, total error $+0.0048\;[+0.0026,+0.0070]$ (all inside $\pm0.02$), and low-precision aligned $s$ $-0.0011\;[-0.0018,-0.0004]$ (inside $\pm0.01$). Acquisition-over-stationary-update total $v$ is $0.8492\;[0.8376,0.8602]$ and $\bar a_v$ ratio is $1.4430\;[1.4233,1.4630]$. In the separate equal-mean weight-history control ($0.80$ constant vs.\ $1.00/0.60$ alternating at fixed balanced acquisition, seeds 9300--9339), own, $m$-swap, $v$-swap, and reset contrasts are $0.0324\;[0.0294,0.0354]$, $0.0321\;[0.0277,0.0362]$, $-0.0141\;[-0.0176,-0.0110]$, and $0.0046\;[0.0031,0.0062]$.

In the fixed-weight follow-up ($w=0.80$), all ten endpoint comparisons in Table~\ref{tab:equalweight} meet their margins. At a common starting multiplier, the within-batch contrast is $-0.0035\;[-0.0083,+0.0015]$ with the original patterns and $+0.0025\;[-0.0021,+0.0071]$ after exchange. Both intervals span zero. Their paired difference (own minus exchanged) is $-0.0060\;[-0.0086,-0.0034]$. The scale target is the geometric mean for each acquisition/update pair. Its mean is $0.6391$ with alternating updates and $0.7304$ with within-batch mixing.

\begin{table}[!htbp]\centering\small
\caption{Endpoint differences with both update routes at $w=0.80$. Each column subtracts the within-batch route from the indicated route (40 paired seeds, mean and 95\% interval).}
\label{tab:equalweight}
\begin{tabular}{@{}lcc@{}}\toprule
Endpoint & Acquisition minus within-batch & Alternating minus within-batch\\
\midrule
Aligned $s$ & $-0.0106\;[-0.0131,-0.0082]$ & $-0.0071\;[-0.0088,-0.0055]$\\
Original-target error & $+0.0074\;[+0.0062,+0.0086]$ & $+0.0060\;[+0.0051,+0.0069]$\\
Destination-target error & $-0.0045\;[-0.0072,-0.0019]$ & $-0.0003\;[-0.0016,+0.0009]$\\
Total error & $+0.0029\;[+0.0006,+0.0051]$ & $+0.0057\;[+0.0050,+0.0064]$\\
Low-precision aligned $s$ & $-0.0024\;[-0.0031,-0.0017]$ & $-0.0019\;[-0.0025,-0.0014]$\\
\bottomrule\end{tabular}
\end{table}

For step count $t$ and parameter subset $\mathcal{S}$ of size $|\mathcal{S}|$, the start-of-pulse mean adaptive scale is
\begin{equation}\bar a_{v,\mathcal{S}}=|\mathcal{S}|^{-1}\sum_{j\in\mathcal{S}}\frac{\eta}{\sqrt{v_j/(1-\beta_2^t)}+\epsilon}.\end{equation}
Global normalization ($\mathcal{S}$ = all $308$ parameters) sets target $a_\ast$ to the geometric mean of the two routes' starting $\bar a_v$ values and rescales $v\leftarrow cv$ via 80 bisection iterations (relative error $<2\times10^{-6}$). Layerwise ($\mathcal{S}\in\{\text{enc1},\text{enc2},\text{head}_A,\text{head}_B\}$), tensor-wise (8 tensors), row-wise ($32$ Layer-1 neurons), and column-wise ($6$ Layer-1 inputs) normalizations apply the same bisection independently within each subset at $t=600$ before continuation begins; subsequent pulse updates evolve $m$ and $v$ under standard Adam. Scalarization sets $v_j=(\eta/a_\ast-\epsilon)^2(1-\beta_2^t)$ across $\mathcal{S}$. The route-specific scalar uses the route's own starting mean multiplier as $a_\ast$. The shared scalar uses the common target $a_\ast$.

On follow-up seeds 20000--20039, calibrated momentum-free SGD uses the per-seed, per-direction geometric mean of first-batch inherited-Adam update-to-regularized-gradient norm ratios as a fixed learning rate ($0.0194$--$0.0529$). At update 10 under inherited Adam, paired route differences in $D_{\mathrm{dest}}$ and $D_{\mathrm{orig}}$ are $-0.1219\;[-0.1263,-0.1174]$ and $+0.1075\;[+0.1032,+0.1114]$, reversing under $v$ swap to $+0.0670\;[+0.0611,+0.0727]$ and $-0.0672\;[-0.0717,-0.0626]$. At update 50, fresh-Adam and calibrated-SGD route contrasts reach $0.0274\;[0.0227,0.0321]$ and $0.0433\;[0.0368,0.0499]$. By update 500, $D_{\mathrm{dest}}$ and $D_{\mathrm{orig}}$ converge to $0.3572$ vs.\ $0.3577$ and $0.9029$ vs.\ $0.9031$. A ten-update pulse followed by neutral continuation likewise shrinks the gap from $0.06197$ at update 50 to $0.00984$ at update 510.

\section{Locating, recording, and perturbing the retained state}
\label{app:mechanism}
\subsection{Multi-granularity scale, layer, and movement controls}
\label{app:layers}
Figure~\ref{fig:scalepattern}a's inherited-Adam result and panel~b's own-state result use the same 40 seed-wise contrasts with separate bootstrap draws. Table~\ref{tab:ivfmechanism} lists both intervals and the remaining plotted estimates.

The per-update movement control runs four branches jointly (acquisition/update $\times$ own/swapped $v$). Each step uses the geometric mean of their four proposed Euclidean lengths, globally or separately in each layer. Each branch keeps its proposed direction within each rescaled group. Adam's moments evolve along the resulting trajectories. Under layerwise equal movement the own and exchanged contrasts are $0.0346$ and $0.0108$.

\begin{table}[!ht]
\centering\small
\caption{Layer exchange, initial-scale, and per-step movement controls on the 40 follow-up seed pairs (seeds 20000--20039). The alternating update route is used unless the row specifies the calibrated within-batch route ($w=0.79$).}
\label{tab:ivfmechanism}
\begin{tabular}{lr}
\toprule
Intervention or paired contrast & Estimate [95\% CI]\\
\midrule
Inherited Adam (Figure~\ref{fig:scalepattern}a) & $0.1907\;[0.1834,0.1973]$ \\
Own state (Figure~\ref{fig:scalepattern}b) & $0.1907\;[0.1837,0.1976]$ \\
Swap first encoder layer $v$ ($5.9\%$ of acq.\ $v$) & $0.0115\;[0.0060,0.0171]$ \\
Swap second encoder layer $v$ & $0.0611\;[0.0558,0.0665]$ \\
Swap both output heads $v$ ($66.2\%$ of acq.\ $v$) & $0.1911\;[0.1838,0.1978]$ \\
Swap all parameters $v$ & $-0.1183\;[-0.1264,-0.1099]$ \\
\addlinespace[2pt]
Route-specific scalar $v$ (own $\bar a_v$) & $0.1251\;[0.1187,0.1318]$ \\
Global common $\bar a_v$: own pattern & $0.0875\;[0.0824,0.0926]$ \\
Global common $\bar a_v$: swapped pattern & $-0.0110\;[-0.0172,-0.0045]$ \\
Global common $\bar a_v$: own minus swapped pattern & $0.0985\;[0.0924,0.1046]$ \\
4-layer equalized $\bar a_v$: own pattern & $0.0875\;[0.0825,0.0923]$ \\
4-layer equalized $\bar a_v$: swapped pattern & $-0.0109\;[-0.0174,-0.0044]$ \\
4-layer equalized $\bar a_v$: own minus swapped pattern & $0.0984\;[0.0925,0.1044]$ \\
8-tensor equalized $\bar a_v$: own minus swapped pattern & $0.0908\;[0.0851,0.0967]$ \\
Layer-1 per-neuron row $\bar a_v$: own minus swapped pattern & $0.0790\;[0.0735,0.0848]$ \\
Layer-1 per-input column $\bar a_v$: own minus swapped pattern & $0.0769\;[0.0718,0.0819]$ \\
Shared scalar $v$ & $0.0395\;[0.0343,0.0449]$ \\
Shared scalar $v$ and zero initial $m$ & $0.0424\;[0.0371,0.0478]$ \\
\addlinespace[2pt]
Within-batch route, global common $\bar a_v$: own pattern & $-0.0065\;[-0.0111,-0.0015]$ \\
Within-batch route, global common $\bar a_v$: own minus swapped pattern & $-0.0060\;[-0.0086,-0.0035]$ \\
Within-batch route, layerwise common $\bar a_v$: own minus swapped pattern & $-0.0037\;[-0.0057,-0.0017]$ \\
\addlinespace[2pt]
Equal global movement, own $v$ & $0.0303\;[0.0269,0.0338]$ \\
Equal global movement, swapped $v$ & $0.0072\;[0.0040,0.0104]$ \\
Equal global movement: paired $v$-swap effect & $0.0231\;[0.0208,0.0253]$ \\
Equal 4-layer movement: paired $v$-swap effect & $0.0238\;[0.0219,0.0256]$ \\
\bottomrule
\end{tabular}
\end{table}

\subsection{Accounting for the formation history}
\label{app:formationlogs}
\label{app:headgrad}
Replaying all 160 follow-up formation branches (40 seeds $\times$ 2 directions $\times$ 2 routes) reproduces every saved endpoint bitwise. Recording unweighted task gradients $g_f,g_o$ before each backward pass decomposes final $v_T$ coordinate-wise as
\begin{equation}v_T=\beta_2^T v_0+S_T+C_T+R_T+E_T,\end{equation}
where $S_T$ accumulates $w_f^2g_f^2+w_o^2g_o^2$, $C_T$ accumulates $2w_fw_og_fg_o$, $R_T$ captures coupled $L_2$ weight decay, and $E_T$ is floating-point remainder; layer cancellation is $-C_T/S_T$. The within-batch follow-up accumulates $S_T$ and $C_T$ during formation (archive IV-G in Appendix~\ref{app:status}).

\begin{figure}[ht]
\centering
\includegraphics[width=.86\textwidth]{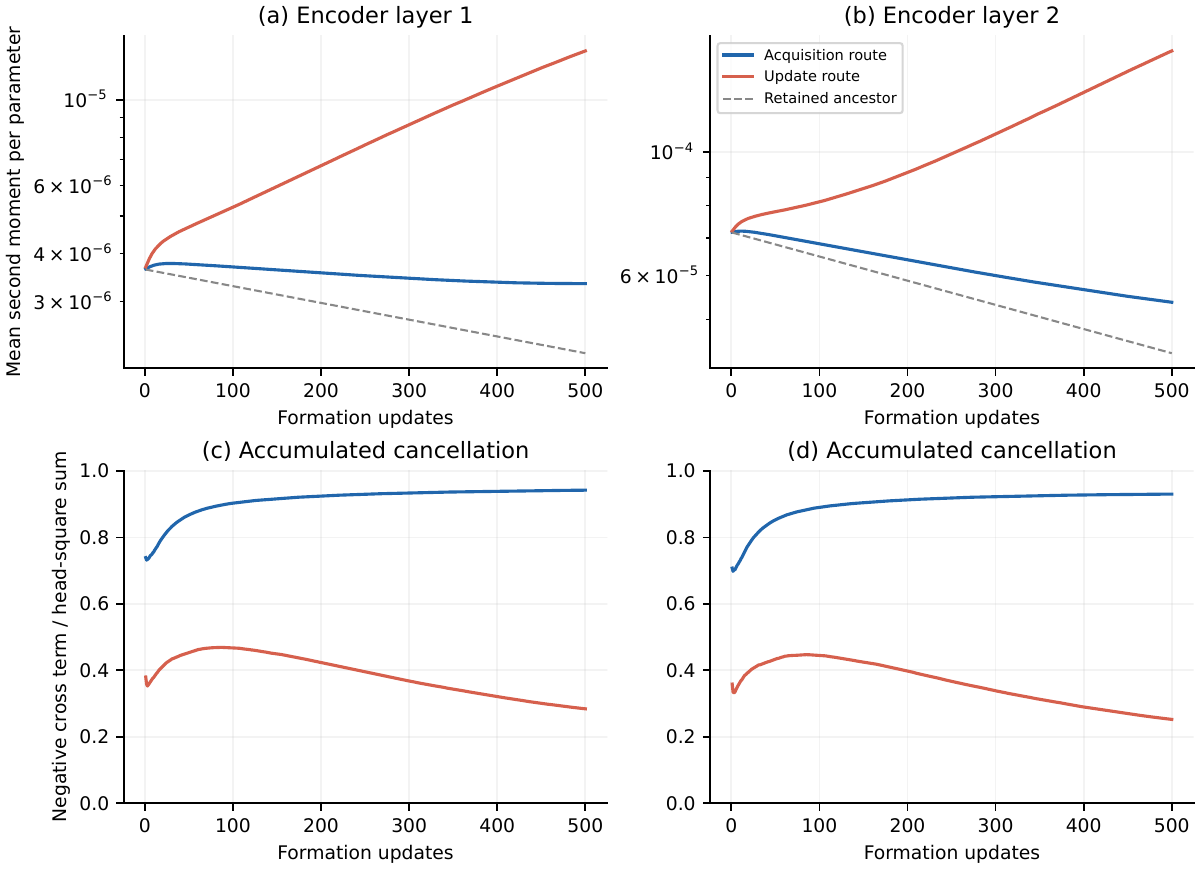}
\caption{Formation-time accounting on the 40 paired follow-up seeds. Top: mean second moment per parameter in each encoder layer; dashed curves show the discounted shared ancestor $\beta_2^t v_0$. Bottom: accumulated cancellation fraction $-C/S$ along each route's formation trajectory. Plotted with Python code developed with OpenAI tools (Section~\ref{sec:ai-methods}).}
\label{fig:formationlogs}
\end{figure}

At update 500, $-C/S$ is $0.942\;[0.939,0.945]$ and $0.930\;[0.927,0.934]$ in encoder layers 1 and 2 under acquisition, versus $0.284\;[0.271,0.298]$ and $0.252\;[0.238,0.268]$ under the alternating update route (Figure~\ref{fig:formationlogs}), and $0.650\;[0.644,0.656]$ and $0.601\;[0.592,0.611]$ under the calibrated within-batch update route ($w=0.79$). Final acquisition/alternating-update $v$ ratios are $0.242\;[0.230,0.255]$ and $0.340\;[0.321,0.361]$, with decayed ancestor $\beta_2^{500}v_0$ supplying $64\%$ and $79\%$ of acquisition final $v$ versus $16\%$ and $28\%$ for alternating update.

Paired acquisition-minus-alternating-update differences in $S_T$ are positive in both encoder layers: $0.000931\;[0.000831,0.001035]$ in layer 1 and $0.000355\;[0.000088,0.000636]$ in layer 2.

Cancellation in the alternating route varies with the target observed precisely. In encoder layer 1, favored-precise steps have mean gradient cosine $-0.81$ and cancellation $0.48$, whereas unfavored-precise steps have cosine $-0.08$ and cancellation $0.06$. In updates 251--500 of the latter condition, the mean cosine is $+0.11$ and cancellation is $-0.12$.

Common-input endpoint gradient probes corroborate these differences: under both-high precision, acquisition endpoints have larger individual head-gradient norms yet a smaller equally weighted sum norm, and first-layer cosines under a favored-precise probe are $-0.940$ versus $-0.835$.

\subsection{Projection followed by crossed state transplantation}
\label{app:projection}
On seeds 38000--38039 (with 37999 reserved for implementation checks), shared-encoder PCGrad replaces conflicting weighted encoder gradients $p,q$ ($p^\top q<0$) by
\begin{equation}p'=p-\frac{p^\top q}{q^\top q}q,\qquad q'=q-\frac{p^\top q}{p^\top p}p,\end{equation}
sending $p'+q'$ to the encoder while keeping head updates and coupled weight decay unchanged.

At formation end, native and projected acquisition recipients each receive native or projected-history $v$ ($2\times2$ design at fixed starting weights, $m$, and $t=600$; 10,000 paired bootstrap resamples, stream 2026101704; Table~\ref{tab:projection}).

\begin{table}[ht]\centering\small
\caption{Crossed second-moment transplants after shared-encoder PCGrad formation: ten-update loss of specialization (40 paired seeds).}
\label{tab:projection}
\begin{tabular}{lcc}\toprule
Recipient model & Native $v$ & Projected-history $v$\\
\midrule
Native acquisition & $0.3551\;[0.3469,0.3634]$ & $0.3079\;[0.3007,0.3154]$\\
Projected acquisition & $0.3435\;[0.3358,0.3512]$ & $0.2948\;[0.2880,0.3016]$\\
\bottomrule\end{tabular}\end{table}

Native acquisition versus update passes all five endpoint bands in this cohort, whereas projected acquisition fails aligned $s$ ($0.0416\;[0.0388,0.0442]$ vs.\ update), $D_{\mathrm{dest}}$, and total error. Projected-own minus native-own ten-update response is $-0.0603\;[-0.0638,-0.0570]$, and raw projected-$v$ donor effects are $-0.0472\;[-0.0489,-0.0455]$ (native recipient) and $-0.0487\;[-0.0503,-0.0470]$ (projected recipient). Rescaling projected-history $v$ at pulse start to the native recipient's global or encoder $\bar a_v$ yields $+0.0113\;[0.0102,0.0123]$ or $+0.0115\;[0.0104,0.0125]$, showing that the negative donor effect changes sign when the initial mean multiplier is matched.

At 50 opposing updates, unscaled projected-history $v$ shifts native and projected recipient responses by $+0.0029\;[0.0025,0.0032]$ and $+0.0021\;[0.0017,0.0026]$, and total projected-own minus native-own contrast reverses to $+0.0440\;[0.0417,0.0461]$.

Exponentially weighted formation traces on the projected trajectory give post/pre projection ratios of $0.117$ for individual squared encoder gradients and $1.844$ for the squared combined gradient.

\section{Regime and target extensions}
\label{app:extension}
The extension studies separate four ten-update claims: endpoint equivalence, positive common-scale pattern effect, unnormalized $v$-swap sign reversal, and reset equivalence within $\pm0.02$.

Nine two-target alternating configurations share seeds 32000--32031: baseline, width 64, ReLU, $\rho=0$, and the $3\times2$ grid of $\beta_2\in\{0.9,0.99,0.999\}$ by $T\in\{500,1500\}$ formation updates (Table~\ref{tab:ivfbeta}).

\begin{table}[ht]
\centering\small
\caption{Decay and formation-duration grid (32 paired seeds per setting). Common-scale pattern effect is own-pattern minus swapped-pattern route contrast at equalized starting $\bar a_v$. All settings pass endpoint equivalence.}
\label{tab:ivfbeta}
\begin{tabular}{llcccc}
\toprule
$\beta_2$ & Formation & Own state & Exchange $v$ & Reset & \shortstack{Common-scale\\pattern effect}\\
\midrule
$0.9$ & 500 & \shortstack{$0.0298$\\$[0.0278,0.0318]$} & \shortstack{$0.0104$\\$[0.0082,0.0127]$} & \shortstack{$0.0373$\\$[0.0342,0.0406]$} & \shortstack{$0.0058$\\$[0.0053,0.0064]$} \\
\addlinespace[3pt]
$0.9$ & 1500 & \shortstack{$0.0764$\\$[0.0736,0.0794]$} & \shortstack{$0.0494$\\$[0.0462,0.0526]$} & \shortstack{$0.1073$\\$[0.1023,0.1123]$} & \shortstack{$0.0120$\\$[0.0111,0.0131]$} \\
\addlinespace[3pt]
$0.99$ & 500 & \shortstack{$0.1385$\\$[0.1329,0.1444]$} & \shortstack{$-0.0720$\\$[-0.0770,-0.0672]$} & \shortstack{$0.0174$\\$[0.0150,0.0197]$} & \shortstack{$0.0673$\\$[0.0625,0.0721]$} \\
\addlinespace[3pt]
$0.99$ & 1500 & \shortstack{$0.2859$\\$[0.2789,0.2925]$} & \shortstack{$0.0297$\\$[0.0204,0.0390]$} & \shortstack{$0.0963$\\$[0.0913,0.1014]$} & \shortstack{$0.1334$\\$[0.1204,0.1448]$} \\
\addlinespace[3pt]
$0.999$ & 500 & \shortstack{$0.1962$\\$[0.1881,0.2043]$} & \shortstack{$-0.1290$\\$[-0.1379,-0.1198]$} & \shortstack{$0.0073$\\$[0.0054,0.0092]$} & \shortstack{$0.1028$\\$[0.0947,0.1109]$} \\
\addlinespace[3pt]
$0.999$ & 1500 & \shortstack{$0.4341$\\$[0.4277,0.4408]$} & \shortstack{$-0.1279$\\$[-0.1416,-0.1139]$} & \shortstack{$0.0574$\\$[0.0529,0.0618]$} & \shortstack{$0.2528$\\$[0.2414,0.2644]$} \\
\bottomrule
\end{tabular}
\end{table}

At $\beta_2=0.9$, retaining optimizer state attenuates the ten-update route contrast at both formation lengths. Resetting Adam increases the contrast by $0.0075\;[0.0059,0.0090]$ after 500 updates and by $0.0308\;[0.0282,0.0334]$ after 1,500 updates.

Across $\beta_2=0.9,0.99,0.999$, paired geometric ratios of the 1,500-to-500-update acquisition/update $\bar a_v$ ratio are $0.792\;[0.759,0.827]$, $0.702\;[0.662,0.747]$, and $1.066\;[1.042,1.091]$.

\subsection{Three Gaussian targets}
\label{app:threeadapt}
Three three-dimensional Gaussian targets with pairwise correlation $0.35$ feed a $9\to32\to d_Z$ $\tanh$ encoder ($d_Z\in\{2,3\}$) and three linear heads for 600 formation updates from a 100-update ancestor.

Pilot seeds 33000--33007 select favored weight $w=0.65$ (others $(1-w)/2=0.175$) at both capacities using endpoints alone; confirmatory seeds 34000--34039 pass all eight endpoint tolerances ($\pm0.02$, or $\pm0.01$ for low-precision indices).

Opposing pulses target destination $j$ at $\sigma=0.05$ with loss weight $0.8$ ($0.1$ on each remaining target) across all six ordered pairs $(i,j)$, measuring $s_{ij}(0)-s_{ij}(t)$ for $s_{ij}=(D_j-D_i)/(D_j+D_i)$.

Both capacities satisfy endpoint matching, $v$-swap reversal, and positive common-scale pattern effects at 10 updates (Table~\ref{tab:targets}); at 50 updates, own-state contrasts are $0.0000/0.0069$, reset contrasts $0.0255/0.0596$, and common-scale effects $-0.0241/-0.0479$.

\subsection{Unequal room-sensor measurements}
\label{app:natural}
The sensor study uses 19,735 chronological ten-minute records from UCI Appliances Energy Prediction \parencite{candanedo2017data,candanedo2017dataset} for kitchen (T1, RH\_1), living room (T2, RH\_2), and laundry room (T3, RH\_3) temperature/humidity pairs.

The first 9,867 rows train the network; four subsequent chronological blocks separated by 144-row gaps fit and evaluate pilot (2,816/2,817 rows) and confirmation (1,829/1,830 rows) readouts, standardizing variables on the training block alone.

Pilot seeds 35000--35007 search $w\in\{1/3,0.40,0.45,\ldots,0.95\}$ per room and capacity ($6\to32\to d_Z$), selecting $0.40/0.95/0.75$ ($d_Z=2$) and $0.95/0.95/0.90$ ($d_Z=3$), yet best candidates still exceed tolerances by $5.30$--$12.44\times$.

Confirmatory seeds 36000--36039 pass only $1/24$ ($d_Z=2$) and $0/24$ ($d_Z=3$) room-specific endpoint criteria, even though direct input readouts confirm precision gains of $0.1784$, $0.4822$, and $0.1824$ across the three rooms; pooled $d_Z=2$ starting differences reach $+0.0616$ in $s_{01}$ and $+0.0238$ at low precision.

At 10 updates, both capacities retain positive common-scale pattern effects (Table~\ref{tab:targets}); at 50 updates, own-state contrasts are $0.0842/0.1156$ and reset contrasts $0.0124/0.0392$.

\begin{table}[ht]\centering\small
\caption{Three-target ten-update contrasts (40 paired seeds). G: Gaussian (matched endpoints); S: UCI room sensors (unmatched endpoints).}
\label{tab:targets}\label{tab:vb}\label{tab:vc}
\begin{tabular}{lcccc}\toprule
Condition & G, $d_Z=2$ & G, $d_Z=3$ & S, $d_Z=2$ & S, $d_Z=3$\\
\midrule
Own & \shortstack{$0.1945$\\$[0.1906,0.1987]$} & \shortstack{$0.4054$\\$[0.3951,0.4152]$} & \shortstack{$0.0601$\\$[0.0456,0.0741]$} & \shortstack{$0.1170$\\$[0.1024,0.1320]$}\\
\addlinespace[3pt]
Exchanged $v$ & \shortstack{$-0.1005$\\$[-0.1054,-0.0955]$} & \shortstack{$-0.2193$\\$[-0.2266,-0.2118]$} & \shortstack{$-0.0254$\\$[-0.0381,-0.0130]$} & \shortstack{$-0.0090$\\$[-0.0232,0.0049]$}\\
\addlinespace[3pt]
Reset & \shortstack{$0.0116$\\$[0.0101,0.0132]$} & \shortstack{$0.0203$\\$[0.0178,0.0226]$} & \shortstack{$-0.0076$\\$[-0.0168,0.0017]$} & \shortstack{$0.0046$\\$[-0.0062,0.0152]$}\\
\addlinespace[3pt]
\shortstack{Common-scale\\pattern effect} & \shortstack{$0.0825$\\$[0.0792,0.0860]$} & \shortstack{$0.1541$\\$[0.1479,0.1597]$} & \shortstack{$0.0103$\\$[0.0064,0.0140]$} & \shortstack{$0.0257$\\$[0.0210,0.0306]$}\\
\bottomrule\end{tabular}\end{table}

\section{Reproduction and evidence status}
\label{app:status}
Time-stamped local plans cover the matched comparison, component exchange, scale interventions, regime extensions and PCGrad. Files named \texttt{preregistration} in the archive are local records. The within-batch comparison at $w=0.79$ uses pilot calibration and a separate 40-seed evaluation. The $w=0.80$ comparison was specified after those outcomes were known and uses the same evaluation seeds. Its analysis plan was fixed before the new runs. Formation logs, checkpoint decompositions and endpoint gradient probes are exploratory.

Archive labels are IV-D (matched responses), IV-E (state exchange), IV-F (network and formation settings), IV-G (within-batch and saved-checkpoint controls), V-B (three Gaussian targets), and V-C (room sensors).

The map \texttt{docs/PAPER\_TO\_DATA.md} links each result to its files in release v1.6.0. The within-batch controls are in \texttt{experiments/iv-g/}.

Confirmatory and follow-up control runs used Python 3.11.7 (formation/projection logging also verified under Python 3.12.14), NumPy 2.4.6, and PyTorch 2.13.0 on single-threaded deterministic CPU, with explicit \texttt{-{}-tolerance 1e-6} cross-environment floating-point checks documented in the README.

\end{document}